\documentclass[letterpaper, 10 pt, conference]{ieeeconf}  

\IEEEoverridecommandlockouts                              
\usepackage{graphicx} 
\usepackage{times}
\usepackage{epsfig} 
\usepackage{mathptmx} 
\usepackage{amsmath} 
\usepackage{amssymb}  
\usepackage{multicol}
\usepackage[bookmarks=true]{hyperref}
\usepackage[usenames]{color}
\usepackage{comment}
\usepackage{float}	
\usepackage{multirow}
\usepackage{multicol}
\usepackage{subcaption}
\usepackage{balance}
\usepackage[export]{adjustbox}
\usepackage[T1]{fontenc}
\usepackage{array}
\usepackage{siunitx}
 
\usepackage{amssymb}
\usepackage{pifont}
\newcommand{\cmark}{\ding{52}}%
\newcommand{\xmark}{\ding{55}}%
\usepackage{booktabs} 
\usepackage{threeparttable} 
\usepackage[ruled,vlined]{algorithm2e}
\usepackage{flushend}
\usepackage{booktabs}
\makeatletter
\let\NAT@parse\undefined
\makeatother
\usepackage[numbers]{natbib}

\title{\LARGE \bf
    In-Hand Cube Reconfiguration: Simplified
}

\author{Sumit Patidar$^{1}$ \quad\quad Adrian Sieler$^{1,2}$ \quad\quad Oliver Brock$^{1,2}$
\thanks{$^{1}$ Robotics and Biology Laboratory, Technische Universität Berlin}
\thanks{$^{2}$ Science of Intelligence, Research Cluster of Excellence, Berlin}%
\thanks{We gratefully acknowledge funding provided by the Deutsche Forschungsgemeinschaft (DFG, German Research Foundation) under Germany’s Excellence Strategy – EXC 2002/1 “Science of Intelligence” – project number 390523135.}
}

\begin{document}

\maketitle
\thispagestyle{empty}
\pagestyle{empty}

\begin{abstract}
  We present a simple approach to in-hand cube reconfiguration. By simplifying planning, control, and perception as much
  as possible, while maintaining robust and general performance, we gain insights into the inherent complexity of
  in-hand cube reconfiguration. We also demonstrate the effectiveness of combining GOFAI-based planning with the
  exploitation of environmental constraints and inherently compliant end-effectors in the context of dexterous
  manipulation. The proposed system outperforms a substantially more complex system for cube reconfiguration based on
  deep learning and accurate physical simulation, contributing arguments to the discussion about what the most promising
  approach to general manipulation might be. \footnotesize{Project website: \url{https://rbo.gitlab-pages.tu-berlin.de/robotics/simpleIHM/}}
\end{abstract}


\section{Introduction}

\begin{figure}[t]
    \centering
    \includegraphics[width=\columnwidth]{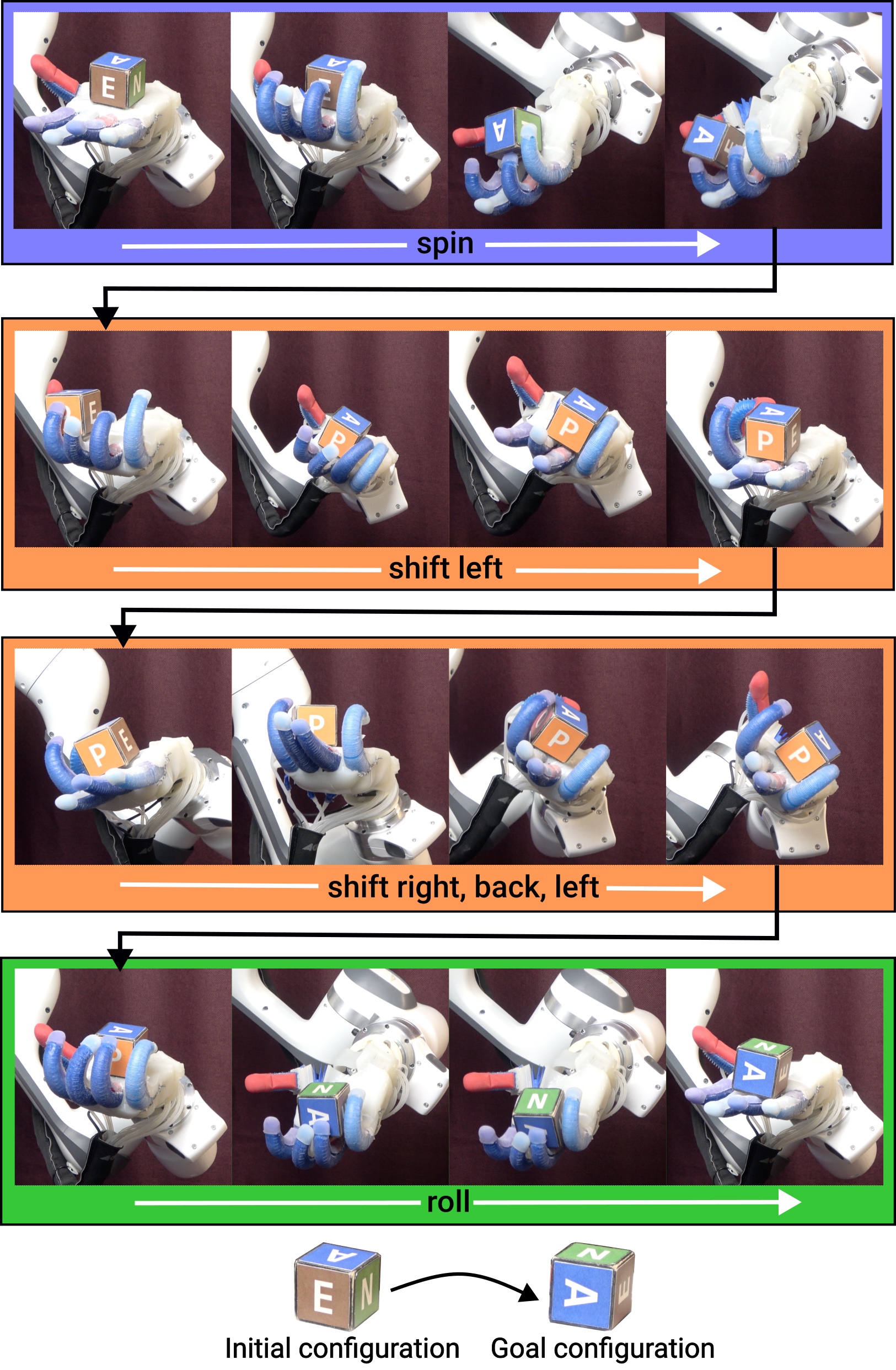}
    \caption{\textbf{A simple and robust approach to in-hand cube reconfiguration:} The cube can be rotated into any of its 24 possible configurations by a sequence of simple and robust motion primitives. These primitives allow to spin (purple), translate (orange) and roll (green) the cube.}
    \label{fig:long_seq}
    \vspace{-0.5cm}
\end{figure}

In-hand cube reconfiguration has recently become a benchmark problem for robot manipulation~\cite{adrian-RSS-21},~\cite{wood_clark},~\cite{morgan_2022} and in particular for the application of deep learning to robot manipulation~\cite{openai_2020},~\cite{MIT},~\cite{DLR_latest},~\cite{touch-dexterity},~\cite{khandate2023sampling}. The objective is to reorient a cube within a robot hand to any of its 24 axis-aligned configurations (see Fig.~\ref{fig:long_seq} for an example).

We investigate the intrinsic complexity of in-hand cube reconfiguration and show that already simple algorithmic tools suffice to solve the problem. This kind of problem analysis, by attempting to develop the simplest possible solution, is inspired by Kolmogorov complexity~\cite{kolmogorov1965three}.
Intuitively, the complexity of generating a desirable behavior corresponds to the smallest number of parameters required to specify a program capable of generating the behavior. Of course, in our context, simplicity cannot be defined in a formal sense. Still, we believe our solution will be considered very simple in terms of modeling and computational costs involved. Concretely, we show that using only gravity as actuation for the manipulandum and exclusively relying on a dexterous, soft hand to produce in-hand environmental constraints~\cite{eppner_ece}, we can robustly and generally solve the in-hand cube reconfiguration problem in an open-loop manner, i.e.~without sensing.

\begin{table*}[ht]
    \vspace*{0.3cm}
    \resizebox{\textwidth}{!}{%
        \begin{threeparttable}[b]
            \centering
            \renewcommand\arraystretch{1.2}
            \caption{\textbf{Comparison of state-of-the-art cube reconfiguring manipulation systems:} The table lists sources of complexity. The approach described in this paper does away with all of them while still solving the cube reconfiguration problem, arguably more robustly than the other methods.}
            \label{tab:rel_work}
            \begin{tabular}{|c|c|c|c|c|c|c|}
                \hline
                \multirow{2}{*}{\textbf{Research work}} & \multirow{2}{*}{\textbf{Robot hand}} & \multirow{2}{*}{\textbf{Symmetry}} & \multirow{2}{*}{\parbox{2cm}{\centering \textbf{Complex control with DL}}} & \multicolumn{2}{c|}{\textbf{Feedback}} & \multirow{2}{2cm}{\textbf{Prehensile manipulation}}          \\
                                                        &                                      &                                    &                                                                            & \textbf{Proprioceptive}                & \textbf{Vision}                                     &        \\
                [0.5ex]
                \hline
                \citet{openai_2020}, OpenAI             & Shadow Dexterous Hand\tnote{1}       & \xmark                             & \cmark                                                                     & \xmark                                 & \cmark                                              & \cmark \\
                \citet{nvidia_dextreme_2022}, Nvidia    & Allegro Hand\tnote{1}                & \xmark                             & \cmark                                                                     & \cmark                                 & \cmark                                              & \cmark \\
                \citet{MIT}, MIT                        & D'Claw Hand                          & \cmark                             & \cmark                                                                     & \cmark                                 & \cmark                                              & \cmark \\
                \citet{morgan_2022}, Yale               & 4 Finger Gripper                     & \cmark                             & \xmark                                                                     & \xmark                                 & \cmark                                              & \cmark \\
                \citet{DLR_latest}, DLR                        & DLR Hand II\tnote{2}                         & \cmark                             & \cmark                                                                     & \cmark                                 & \xmark                                              & \cmark \\
                \citet{touch-dexterity}, HKUST & Allegro Hand\tnote{1} & \xmark & \cmark & \cmark & \xmark & \cmark \\
                \citet{khandate2023sampling}, Columbia & 5 Finger Gripper & \cmark & \cmark & \cmark & \xmark & \cmark \\
                \citet{adrian-RSS-21}, TUB              & RBO Hand 3\tnote{1}                  & \xmark                             & \xmark                                                                     & \xmark                                 & \xmark                                              & \cmark \\
                \hline
                \textbf{Ours}                           & RBO Hand 3\tnote{1}                  & \xmark                             & \xmark                                                                     & \xmark                                 & \xmark                                              & \xmark \\
                [0.5ex]
                \hline
            \end{tabular}
            \begin{tablenotes}
                \item [1] General purpose, dexterous, anthropomorphic hand
                \item[2] General purpose, dexterous, anthropomorphic hand with a particular symmetric finger configuration
            \end{tablenotes}
        \end{threeparttable}
    }
\end{table*}

We demonstrate the viability of defining a small, robust, and simple set of in-hand manipulation primitives performed on a general-purpose soft anthropomorphic hand to enable simple planning techniques for solving the cube reconfiguration problem. The primitives consist of wrist motions to change the direction of gravity with respect to the hand/object system. Gravity actuates the object to perform a motion constrained by the hand's morphology to achieve robustness and generality. We show experimentally that cube reconfiguration is possible with high success rates based on simple control and planning. We also compare our results to the performance of OpenAI's cube reconfiguration~\cite{openai_2020}, a deep learning approach based on a neural network with more than one million parameters, requiring days of simulation time, sophisticated sensing, and presumably consuming energy roughly equivalent to hourly output of three nuclear power plants~\cite{wired_ai}.

Our results inform the quest for a solution for general robot manipulation. The results support the suitability of GOFAI planning methods in manipulation---if these planning methods can rely on a small set of robust primitives that serve as a basis for diverse behaviors. Such primitives, we believe, can be designed or learned and must be based on the exploitation of environmental constraints~\cite{eppner_ece} as well as the robustness contributed by inherently compliant (soft) end-effectors~\cite{adrian-RSS-21}. The future will tell if such a GOFAI-based approach coupled with insights from soft robotics can prove sufficiently powerful to address general manipulation. Given state of the art, however, Occam's razor might favor this approach over much more complex alternatives.

\section{Related Work}


\subsection{Cube Reconfiguration}

Recently, several manipulation systems have been designed to solve cube reconfiguration tasks. Most of these works use deep reinforcement learning to learn policies for dexterous cube manipulation~\cite{openai_2020},~\cite{MIT},~\cite{DLR_latest},~\cite{touch-dexterity},~\cite{khandate2023sampling},~\cite{nvidia_dextreme_2022}. We would classify these systems as complex based on various measures of complexity: They require a lot of computation time, have a very large number of parameters, require sophisticated sensing, and depend on complex and accurate dynamic simulators, to name a few.

Approaches we would classify as simpler rely on designed motion primitives. ~\citet{adrian-RSS-21} also use motion primitives for planning, sequencing them in an open-loop manner. This approach is probably the simplest approach found in the literature. \citet{morgan_2022}, in addition to primitives, uses object pose feedback at smaller time steps for executing the primitives, adding some complexity.

Interestingly, only few researchers~\cite{adrian-RSS-21},~\cite{openai_2020},~\cite{nvidia_dextreme_2022}, solve the cube reconfiguration task on a general-purpose, anthropomorphic hand. In these approaches, the palm of the hand is facing upwards. In contrast, others manipulate the cube with the palm facing downwards~\cite{morgan_2022},~\cite{MIT},~\cite{DLR_latest} but not with general-purpose hands. Instead, they rely on hands in which the fingers are arranged symmetrically. This particular arrangement simplifies the learning problem but will prove disadvantageous for other manipulation tasks. The need to counteract the effects of gravity also necessitates object-level sensing in these approaches.

So far, none of the approaches leverage the full potential of wrist movements to generate robust dexterous behavior. They all rely on the intrinsic degrees of freedom the hand offers to solve in-hand manipulation tasks. Therefore, all these works require, in contrast to our work, an object to be firmly grasped.

We summarize the cube manipulation works in Table~\ref{tab:rel_work}. Symmetry refers to the reliance of a system on a particular symmetric finger configuration. We view this symmetric arrangement as a form of complexity as the finger configuration tends to be carefully chosen in these examples. Using deep learning (DL) for control and perception also adds complexity, as it requires data, training, simulation, and computation. Clearly, not requiring perception simplifies the system. And without perception, only open-loop control is possible, a simple form of control. Finally, prehensile manipulation requires achieving and maintaining force-closure hand/object configurations, also adding complexity that is not required for non-prehensile manipulation. None of these sources of complexity are present in the cube reconfiguration system described in this paper.

\subsection{Leveraging Physical Constraints, Gravity, and Inertia}

~\citet{eppner_ece} describe ``Environmental Constraint Exploitation'' (ECE) as a way for robots to manipulate objects by deliberately using contacts with the environment. These deliberate contacts with the environment restrict object motion and allow grasping and repositioning objects in the hand~\cite{eppner_ece},~\cite{dafle_extrinsic_2014},~\cite{dafle_prehensile_pushing}. In this work, the hand itself provides such compliant constraints~\cite{steffan_rh3_22}. We configure these constraints (by reconfiguring the hand) to form partial cages~\cite{partial_caging} that restrict the mobility of an object to a portion of configuration space and thus reduce uncertainty in object pose without sensing.

In~\cite{dafle_extrinsic_2014},~\cite{erdmann_sensorless_1988},~\cite{erdmann_nonprehensile_1998}
gravity and inertia are exploited to manipulate objects using simple grippers. Similarly, we exploit compliant constraints provided by a general-purpose robot hand using gravitational and interial forces to move an object within the hand.


\section{Formulating Motion Primitives}

We design motion primitives by the simple interplay of physical constraints and
gravity on a soft dexterous hand. Before we discuss the details of these motion
primitives, we introduce the robotic setup in the following section.

\subsection{The Robotic Platform}\label{sec:hardware_setup}

We use the soft, anthropomorphic RBO~Hand~3~\cite{steffan_rh3_22} which has 16 degrees of actuation (DoA) as the
end-effector. The hand consists of soft continuum and pouch actuators which are pneumatically actuated and are
inherently compliant. We actuate the hand by controlling the air mass enclosed in each actuator.

\begin{figure}[h]
    \centering
    \begin{subfigure}[b]{0.22\textwidth}
        \centering
        \includegraphics[width=\textwidth]{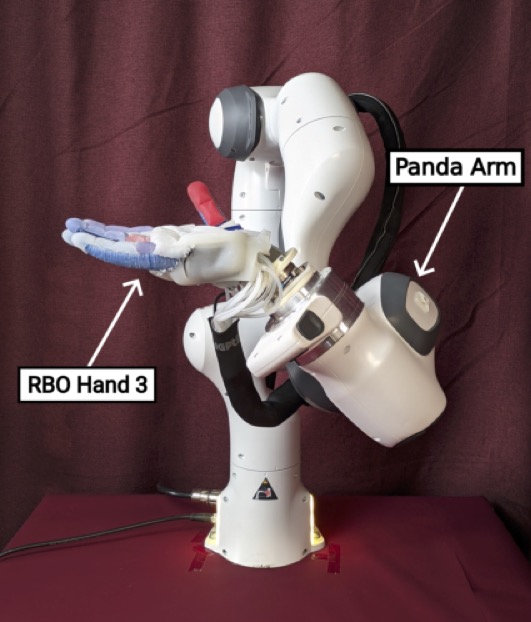}
        \label{fig:hardware_setup}
    \end{subfigure}
    \begin{subfigure}[b]{0.22\textwidth}
        \centering
        \includegraphics[width=0.7\textwidth]{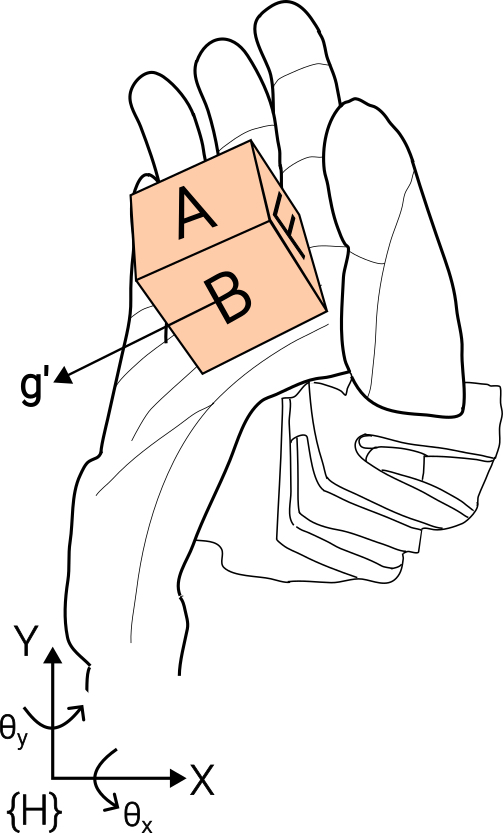}
        \label{fig:gv}
    \end{subfigure}
    \caption{\textbf{Robot setup} consisting of the RBO Hand 3 attached to the Franka Emika Panda arm. Gravity is controlled by reorienting the wrist.}
    \label{fig:setup_and_control}
\end{figure}

The hand is attached to a Franka Emika Panda robot arm as shown in Fig.~\ref{fig:setup_and_control}. We use the arm to
leverage gravity by reorienting the hand, as shown in Fig.~\ref{fig:setup_and_control}. We also wiggle the hand to
leverage the inertial effects of an object to trigger object motion by overcoming static friction between object and
hand.

\subsection{Using Gravity and Inertia for Actuation}

The manipulated object is always supported by the palm and not in a force-closure grasp. The hand, with its many
actuated degrees of freedom, is reconfigured to provide diverse sets of environmental constraints. Each set can be
interpreted as a partial cage that the object can move around in. The fingers are not used to actively exert forces onto
the object. Gravity and inertia are the only active actuation sources. We control the wrist orientation to leverage
gravity and wiggle the hand to make sliding easy on the high-friction hand surface. By tilting the wrist around the $x$-
and $y$-axis, we can span the effective gravity vector across the palm plane of the hand, as described by
Eq.~\ref{eq:tilt_from_gravity}.

\begin{align}
    \begin{bmatrix}
        \theta_x \\ \theta_y
    \end{bmatrix} & = \arcsin(\alpha) \begin{bmatrix}
                                          -\hat{g'}_y \\ \hat{g'}_x
                                      \end{bmatrix}.
    \label{eq:tilt_from_gravity}
\end{align}

$\theta_x$ and $\theta_y$ represent the wrist orientation about the $x$- and $y$-axis of the hand frame $H$. $\hat{g}_x$ and $\hat{g}_y$ represent the $x$- and $y$-components of the unit effective gravity vector in hand frame $H$. $\alpha$
represent the scale of the gravity magnitude ($9.8~\unit{m.s^{-2}}$) in the range [0,1]. Unlike related work~\cite{bai2014dexterous}, which uses an analytic approach to find tilting angles to move the object, we derive these experimentally for each primitive. In our case, the tilting angles remain constant irrespective of the object to be manipulated.

Additionally, we do not model inertial forces, friction, and their interactions. Instead, we simply leverage them by wiggling the hand. For wiggling, we oscillate the wrist with an amplitude of $5^\circ$ and frequency of \num{5} \unit{\hertz} relative to the current robot pose. In the next section, we discuss the interplay of constraints and wrist movements to derive motion primitives.

\subsection{Constraint Exploitation using Gravity}\label{sec:primitives_planning}

\begin{figure}[H]
    \vspace{0.25cm}
    \begin{center}
        \includegraphics[width=0.85\columnwidth]{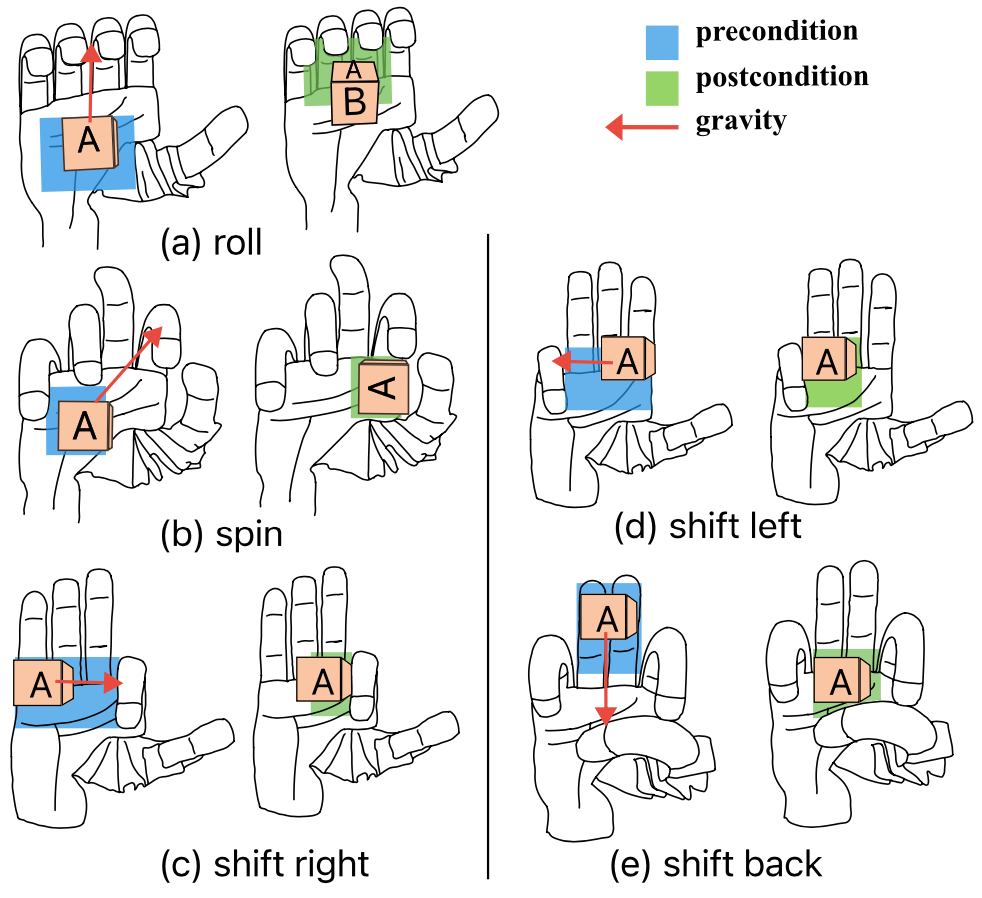}
    \end{center}
    \caption{\textbf{Simple set of motion primitives} to manipulate a cube by exploiting constraints provided by the hand and leveraging gravity as the actuation source.}
    \label{fig:skills}
\end{figure}

To reconfigure a cube to any of its 24~configurations, we derive a set of five simple motion primitives as shown in Fig. \ref{fig:skills}. For each motion primitive, we first create a specific constraint arrangement, and then exploit it by moving the wrist. The roll (Fig.~\ref{fig:skills}a) and spin (Fig.~\ref{fig:skills}b) primitives rotate the cube along two orthogonal axes. Both rotations change not only the orientation of the cube but also its position. Therefore, if we want to sequence these primitives, we need to restore the cube to its original position before the rotation. We derive such primitives by sliding the cube against a finger or a group of fingers that act as a planar environmental constraint (Fig. \ref{fig:skills}c-e). Each primitive has a pre- and postcondition set that are used for planning. The precondition set describes the positions of the object where the primitive is likely to succeed. The postcondition set describes the set of configurations of the object after the primitive is executed. These sets are required for GOFAI planning. In addition, each primitive requires two inputs: the actuation levels of the hand to achieve the constraint configuration, and the corresponding wrist orientation to induce object motion via gravity as described in Routine~\ref{algo:primitive}.

We hand-craft the primitives based on our intuition of the constraints and wrist motion needed to achieve the desired object motion. We experimentally derive these primitives using a wooden cube of size \num{4.7} \unit{\cm} and weight of \num{70} \unit{\gram}. In a given constraint configuration, we determine the direction of gravity and then adjust the wrist orientation until the cube slides or rolls according to the desired motion. If the cube does not move due to irregularities on the palm surface or high friction, we wiggle the hand after tilting.

To reconfigure a cube to any of its 24 configurations, we sequence these primitives based on their effect on the cube. By considering the initial arrangement of letters on the cube and the pre- and postconditions of each primitive, we utilize breadth-first search to create a plan that brings the cube from its starting configuration to the desired goal configuration. This plan consists of spin and roll rotations and shift operations to reset the cube's position for subsequent rotations.

\begin{algorithm}
    \SetAlgorithmName{Routine}{}

    \SetKwFunction{ConfigureConstraints}{ConfigureConstraints}
    \SetKwFunction{ResetWrist}{ResetWrist}
    \SetKwFunction{ResetHand}{ResetHand}
    \SetKwFunction{ReorientWrist}{ReorientWrist}
    \SetKwFunction{OscillateWrist}{OscillateWrist}
    \SetKwInOut{Input}{input}\SetKwInOut{Output}{output}

    \Input{$air\_mass$, $\theta$, wiggle = {true or false}}
    \Output{manipulation behavior that translates or rotates the cube}
    \BlankLine
    {

        \ConfigureConstraints{air\_mass}\ \tcp*[r]{actuate hand}
        \ReorientWrist{$\theta$}\  \tcp*[r]{gravity}
        \If{wiggle}{\OscillateWrist{}\;
        }
        \ResetWrist{}\ \tcp*[r]{home pose}
        \ResetHand{}\ \tcp*[r]{deflate hand}
    }
    \caption{Motion Primitive}
    \label{algo:primitive}
\end{algorithm}

\section{Solving Cube Reconfiguration Task}

To experimentally assess the robustness of our hand-crafted motion primitives, we compute the success rate for each primitive when applied to different objects. Afterwards, we generate long manipulation plans to visit multiple cube configurations and compare the performance to a deep learning based solution. The robotic platform we use is the one introduced in~Sec.~\ref{sec:hardware_setup} consisting of RBO Hand 3 mounted on a Franka Emika Panda. 

\subsection{Robustness of Motion Primitives}

\begin{figure}[h]
    \vspace*{0.25cm}
    \centering
    \includegraphics[width=0.9\columnwidth]{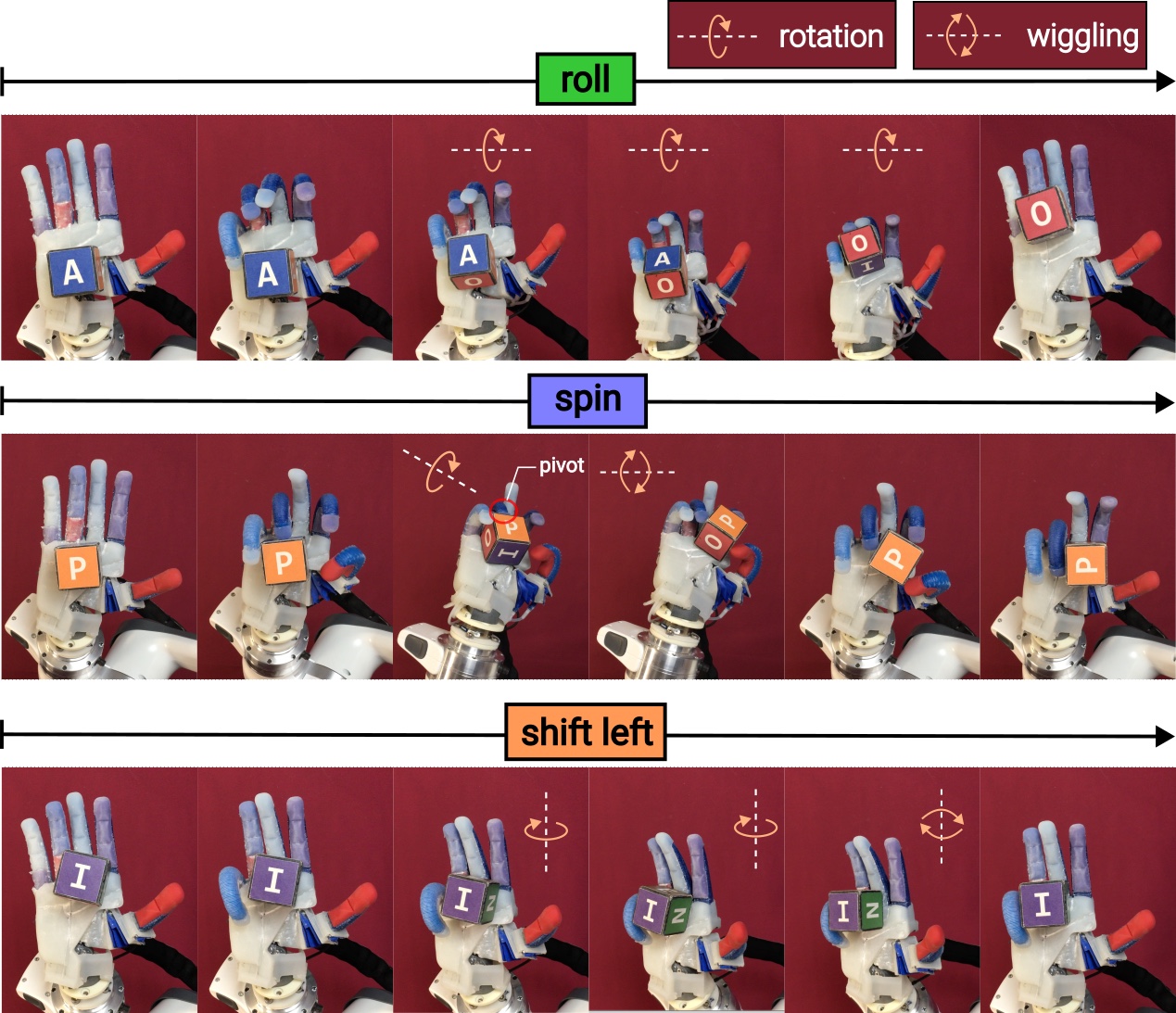}
\caption{\textbf{Keyframes of the roll, spin, and shift (left) primitives} while execution on the real robot (top view).}
    \label{fig:motion_primitives}
    \vspace{-0.5cm}
\end{figure}

\begin{figure}[t]
    \vspace*{0.25cm}
    \centering
    \includegraphics[width=0.9\columnwidth]{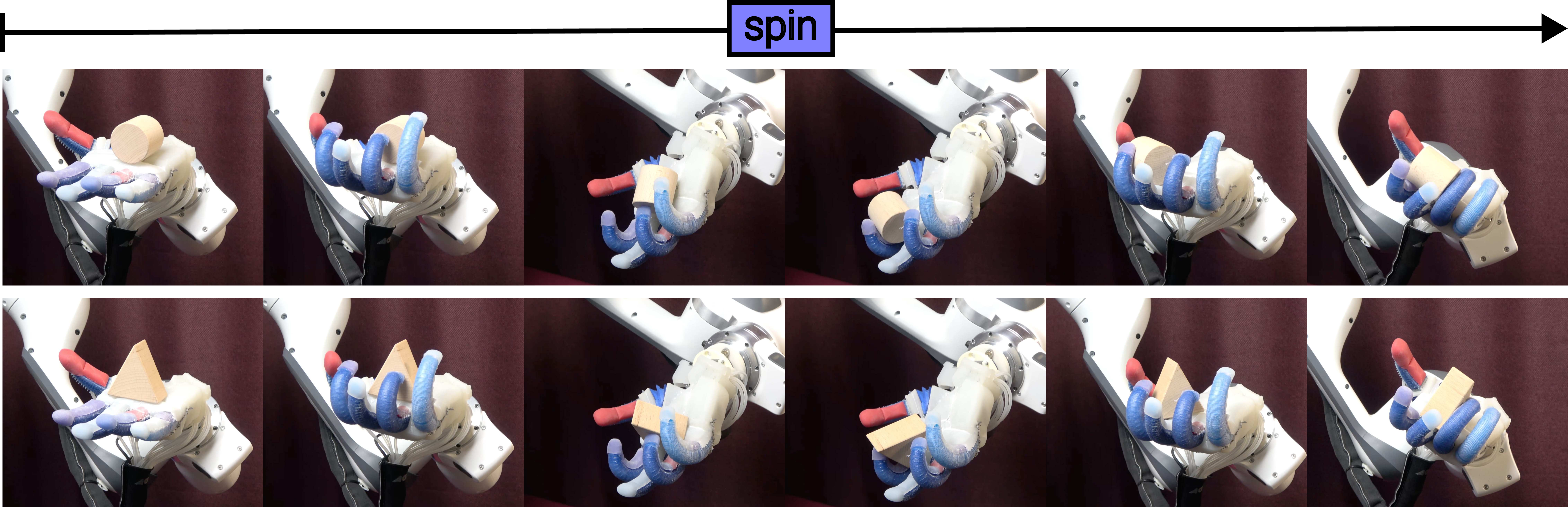}
    \caption{\textbf{Keyframes of the spin primitive for prism and cylinder (side view).} The prism's lateral and the cylinder's curved side are supported by the palm.}
    \label{fig:spin_primitive_on_cylinder_and_prism}
    \vspace{-0.5cm}
\end{figure}

For each primitive, we place the cube initially at a randomly chosen position from its precondition region. Then, we
execute that motion primitive multiple times and observe the manipulation behavior. For each trial, if the cube's pose
changes in a desired manner, i.e. translates (for translation primitives) or rotates (rotation primitives), we count the
trial as a success, otherwise as a failure. One such trial is depicted in Fig.~\ref{fig:motion_primitives}. The results
for 20~trials for each motion primitives and three different objects are given in table~\ref{tab:mot_prim}. The motion
primitives in our work have a high success rate and generalize across variations in the initial cube placements spanned
across the precondition set.
The overall behavior produced by these primitives is robust, even though they are generated with simple open-loop
actuation. We use five different objects to test derived motion primitives (Tab. \ref{tab:objects_used}), and to our
surprise, these generalize well to different shapes (Fig. \ref{fig:spin_primitive_on_cylinder_and_prism}) given the same
actuation and pre- and postcondition set. 
Consequently, they can be sequenced into more complex object manipulations. In the following section, we investigate the robustness of
sequences of these primitives to solve the cube reconfiguration tasks.

\begin{table}[t]
    \vspace{0.25cm}
    \centering
    \small
    \caption{\textbf{Objects used in the experiments}. We use cube, cuboid, cylinder, and prism objects for
    manipulation.}
    \label{tab:objects_used}
        \begin{tabular}{c c c}
            \hline
            \textbf{Object}  & \textbf{Dimensions (\unit{\cm})}   & \textbf{Weight (\unit{\gram}})            \\
            [0.5ex]
            \hline
            Cube $\mathrm{I}$ \includegraphics[width=8pt]{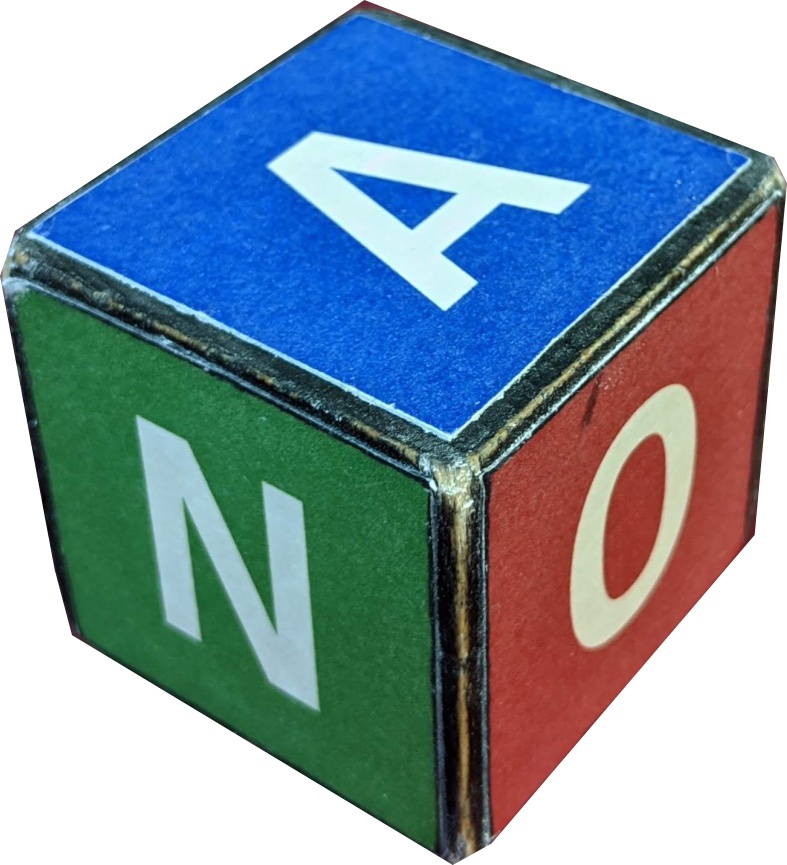}  & 4.7 & 70 \\
            Cube $\mathrm{II}$ (Rubik's)\includegraphics[width=8pt]{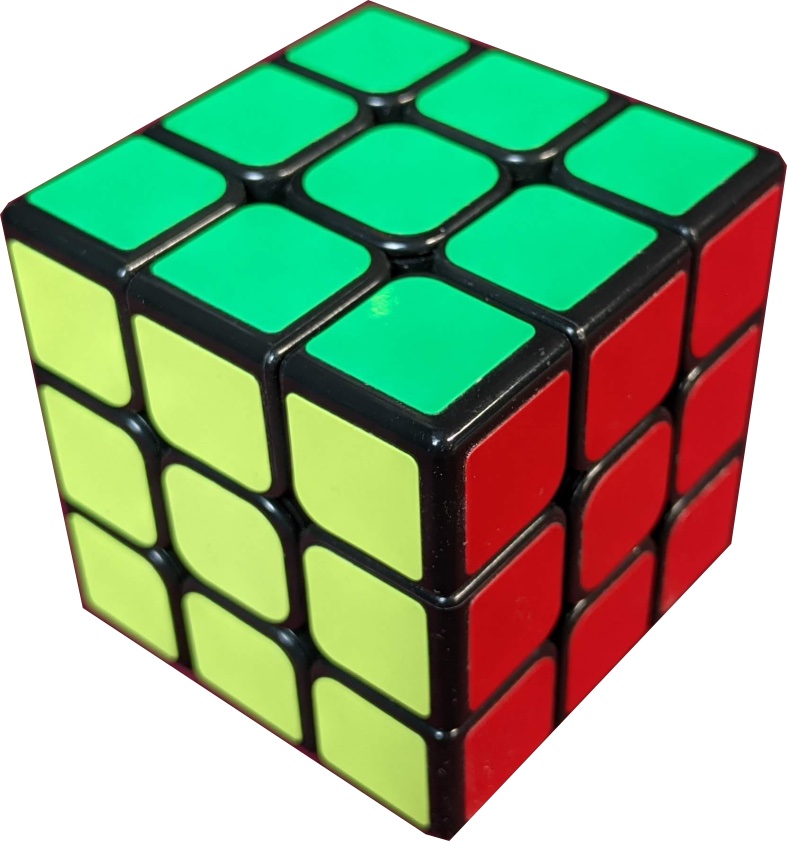} & 5.7 & 78 \\
            Cuboid \includegraphics[width=8pt]{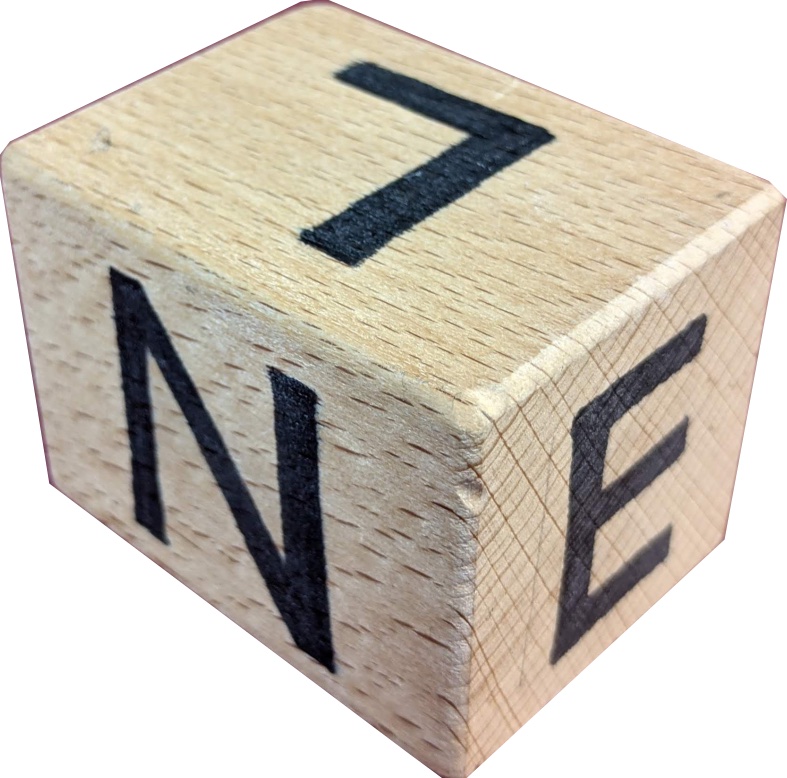} & \numproduct{6 x 4.5 x 4.5}  & 85\\
            Cylinder \includegraphics[width=8pt]{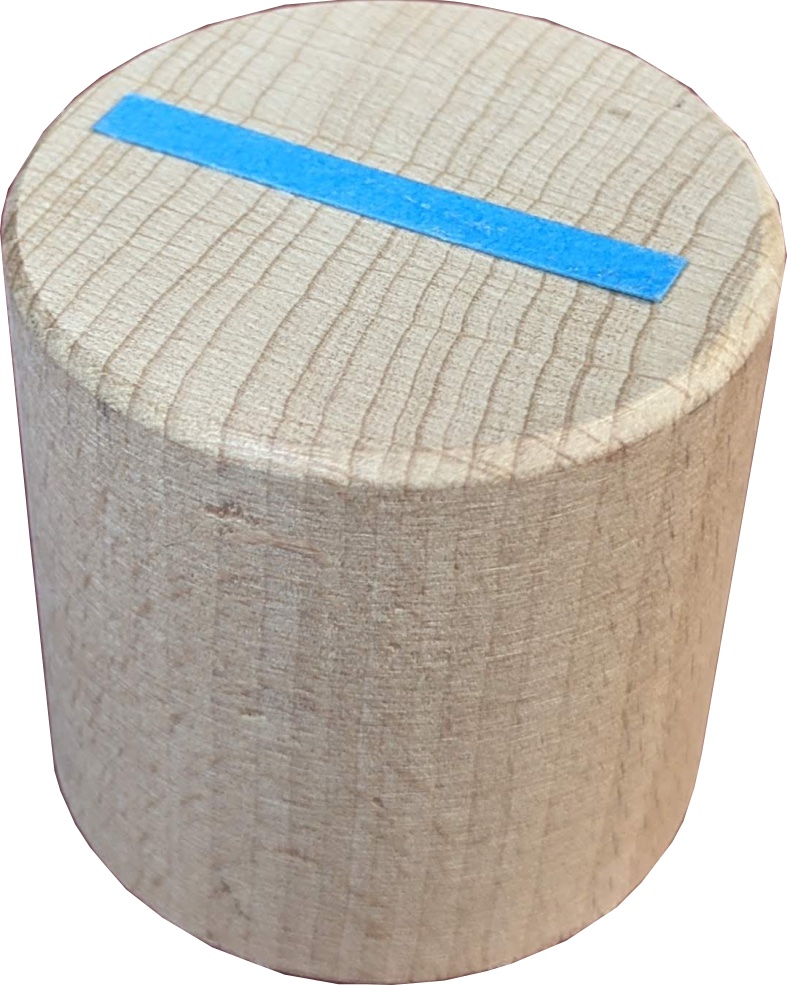}& r = 2.2, h = 4.8 & 53 \\
            Prism \includegraphics[width=8pt]{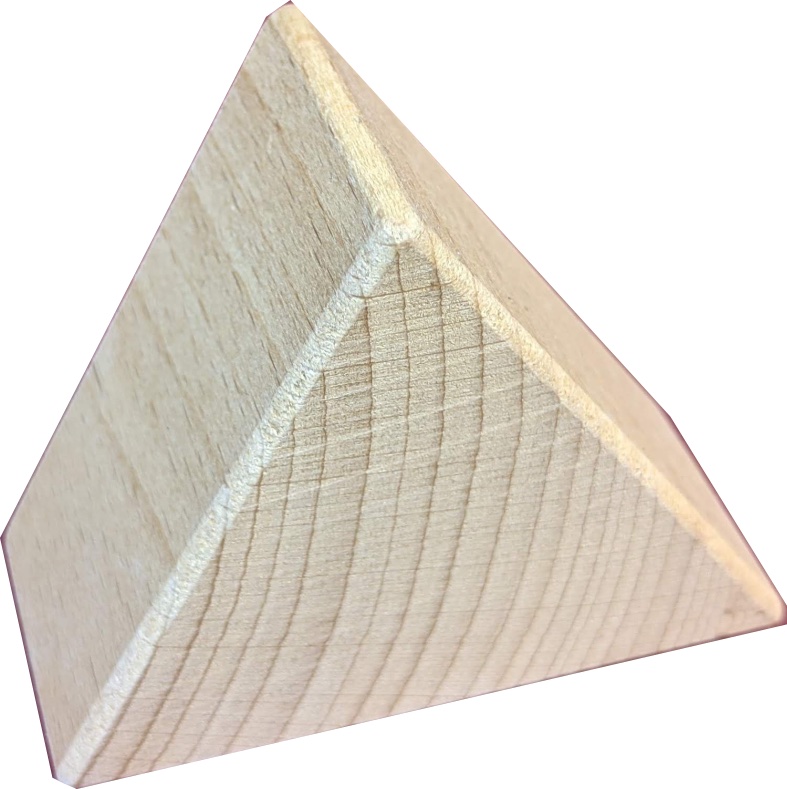}& \numproduct{6 x 5.7 x 3} & 35 \\
            [0.5ex]
            \hline
        \end{tabular}
        \vspace{-0.25cm}
\end{table}

\begin{table}[h]
    \small
    \centering
    \renewcommand\arraystretch{1.2}
    \caption{\textbf{Generalization of motion primitives to object variations.} Successful executions of motion primitives over 20 trials each on the physical robot: If the cube rotates by $90^\circ$ for roll and spin, it is counted as a success. For different shift variations (right, left, and back), if the cube translates fully in the desired direction and aligns with the target constraint, it counts as success; otherwise as a failure.}
    \label{tab:mot_prim}
    \begin{tabular}{lccc}
        \hline
    \multirow{2}{*}{\textbf{Motion Primitives}} & \multicolumn{3}{c}{\textbf{Successful executions}}                                                                                                                                                       \\
                                                & Cube $\mathrm{I}$ & Prism & Cylinder \\
                                                [0.5ex]
                                            \hline

    Roll (R)                                    & 20/20                                                       & 20/20                                                            & 20/20                                                                   \\
    Spin (S)                                    & 20/20                                                       & 12/20                                                            & 15/20                                                                   \\
    Shift right ($T_r$)                          & 20/20                                                       & 10/20                                                            & 20/20                                                                   \\
    Shift left ($T_l$)                           & 20/20                                                       & 10/20                                                            & 20/20                                                                   \\
    Shift back ($T_b$)                           & 18/20                                                       & 8/20                                                             & 20/20   \\                                                               
    \hline
    \end{tabular}
\end{table}

\subsection{Long Composition of Motion Primitives}

We have shown that the motion primitives for the wooden cube are highly reliable. Therefore, the cube reconfiguration
problem becomes a simple planning problem because only a small set of primitives is required and no feedback is needed.
We rotate a cube to its different faces by sequences of the five derived motion primitives that are derived with
breadth-first search as described in Sec.~\ref{sec:primitives_planning}. The experimental data of executing these plans
is reported in Table~\ref{tab:long_seq}. On our \href{https://rbo.gitlab-pages.tu-berlin.de/robotics/simpleIHM/}{project page}, long sequence demo video shows execution of 34 sequenced
primitives on the wooden cube, while the spelling demo video involves two different cube sizes and a
cuboid, executing in total 21 primitives across all objects.


We compare our results with the performance achieved by OpenAI's cube reconfiguration system (Tables~\ref{tab:long_seq}
and~\ref{tab:openai}). We do only have three trials as OpenAI evaluated their long manipulation sequences ten times. This is the case because we did not start this research with the intention of comparing it to OpenAI. Instead, we focused on building a simple manipulation system relying on inertial and gravitational forces for IHM without any perception. The observed robust behavior gave rise to conduct this comparison. We compare OpenAI's total subsequent rotations vs. subsequent total primitives (rotation + translations) because all our individual primitives move the object within the hand while moving the RBO Hand 3 and the Franka Panda. First, our simpler method reliably reconfigures the cube using long sequences (Exp. 4, 7, 8 in Table \ref{tab:long_seq}) with almost zero standard deviation compared to OpenAI's system which exhibits high variance. 
Second, OpenAI's system performs worse if they use wrist movements during manipulation. In contrast, we generate robust
behavior by purely relying on the wrist as the actuation source without any feedback. Third, the manipulation behavior produced by our system is easily legible by humans (facilitating human/robot collaboration) and does not include the seemingly random, unexplainable, not goal-oriented finger motions present in OpenAI's system.


Our findings reveal that in-hand cube reconfiguration is a relatively simple manipulation problem, indicated by the
simplicity of the presented system. Therefore, cube reconfiguration may not be a good benchmark to demonstrate the
effectiveness of deep learning for manipulation tasks. Furthermore, our results show that very simple solutions can
compete with deep learning methods for cube manipulation.

\begin{table}
    \vspace*{0.25cm}
    \centering
    \renewcommand\arraystretch{1.2}
    \caption{\textbf{Successful executions of cube reconfiguration plans} to reach different goal configurations (three trials for each plan). We count the successful consecutive executions of motion primitives until the first failure of any primitive is observed. Here, R \& S represent the roll and spin primitive, and T represents the combination of three shift primitives $[T_l, T_r, T_b]$. $N_R, N_T$  represent the number of rotation and translational primitives, respectively. $\mathbf{N}$ is the total number of primitives in the plan. The column ``individual trials'' lists the number of consecutive primitives executed successfully in each trial (maximum: $\mathbf{N}$). Mean values for successful consecutive primitives are reported in absolute terms and as a percentage of $\mathbf{N}$. All plans are executed on Cube~$\mathrm{I}$.}
    \label{tab:long_seq}
    \resizebox{\columnwidth}{!}{%
        \begin{tabular}{cccccccc}
    
            \hline
            \textbf{Exp.} & \textbf{Plan}         & \textbf{$N_T$} & \textbf{$N_R$} & \textbf{N} & \textbf{Mean $\mathbf{\pm}$ s.d.} & \textbf{Mean (\%)} & \textbf{Individual trials} \\
            [0.5ex]
            \hline
            1             & $(RT)^2$              & 6              & 2              & 8   & 7.33 $\pm$ 0.58  &   91.6   & 8, 7, 7                    \\
            2             & $(RT)^3$              & 9              & 3              & 12 & 11.33 $\pm$ 0.58  &   94.4   & 12, 11, 11                   \\
            3             & $(ST)^2$              & 6              & 2              & 8 & 8 $\pm$ 0   &   100      & 8, 8, 8                    \\
            4             & $(ST)^{10}$           & 30             & 10             & 40 & 40 $\pm$ 0  &   100       & 40, 40, 40                 \\
            5             & $ST,RT,ST$            & 9              & 3              & 12 & 12 $\pm$ 0  &   100       & 12, 12, 12                 \\
            6             & $(RT)^2,ST$           & 9              & 3              & 12 & 10.33 $\pm$ 2.29  &   85.8 & 12, 12, 7                 \\
            7             & $(ST)^2,RT,(ST)^2,RT$ & 18             & 6              & 24 & 23.33 $\pm$ 0.58  &   97.2       & 24, 23, 23                 \\
            8             & $(ST)^3,RT,(ST)^3$    & 21             & 7              & 28 & 28 $\pm$ 0  &   100       & 28, 28, 28                 \\
            [0.5ex]
            \hline
        \end{tabular}
    }
\end{table}

\begin{table}
    \large
    \centering
    \renewcommand\arraystretch{1.2}
    \caption{\textbf{OpenAI results~\cite{openai_2020}}: The number of successful consecutive cube rotations (maximum:
    \textbf{N}) on the real robot. For explanation see Table~\ref{tab:long_seq}.}
    \label{tab:openai}
    \resizebox{\columnwidth}{!}{%
        \begin{tabular}{c c c c c}
            \hline
            \textbf{Physical Task}  & \textbf{N}  & \textbf{Mean $\mathbf{\pm}$ s.d.}   &  \textbf{Mean (\%)} & \textbf{Individual trials}            \\
            [0.5ex]
            \hline
            Block (state)            & 50 & 18.8 $\pm$ 17.1 & 37.6 & 50, 41, 29, 27, 14, 12, 6, 4, 4, 1    \\
            Block (state, locked wrist) & 50 & 26.4 $\pm$ 13.4 & 52.8 & 50, 43, 32, 29, 29, 28, 19, 13, 12, 9 \\
            Block (vision)              & 50 & 15.2 $\pm$ 14.3 & 30.4 & 46, 28, 26, 15, 13, 10, 8, 3, 2, 1    \\
            [0.5ex]
            \hline
        \end{tabular}
    }
\end{table}


\subsection{Limitations and Future Work}

Since the object is not firmly grasped, our approach cannot be applied to upside-down manipulation tasks. The primitives work by partially caging the object. Therefore, large objects that don't completely fit in the hand are hard to manipulate. Additionally, the planning is open-loop, therefore, is not robust to strong external disturbances. However, we can incrementally increase the complexity by including object perception and by learning the parameters of the motion primitives (tilt angle, air mass). We can use RL to sample and learn these small sets of parameters for diverse objects. This would be computationally less expensive and could potentially be carried out in the real world.

\section{Simplicity and Complexity}

There is no agreed-upon definition of simplicity (or complexity) for robotic systems and algorithms, as in the case of
Kolmogorov complexity. We, therefore, have to rely on intuition. Complicating matters further, there is ambiguity in
what aspects of a system we want to consider regarding complexity. Is a neural network complex simply because it has
many parameters? Probably not---if we have a simple way of determining these many parameters, we might not care about
their number. If, however, finding these parameters requires significant engineering effort (simulators) and resources
(computation time, data), we might say that this constitutes some form of complexity.

In regards to \textit{perception}, we can say that---everything else being equal---a system without perception is
simpler than a system with perception. In that sense, the system we propose here is simple. The proposed system uses breadth-first \textit{planning} on a fully known state representation based on a set of five
manipulation primitives. This is simple algorithmically and also in terms of computation time. There is, of course,
exponential growth in computation time for increasing the size of the state space and number of primitives. However, we
believe that both of these numbers remain tractable, even for complex in-hand manipulation problems.

Based on our experiences with the RBO~Hand~3, we designed a set of only five \textit{primitives}, each simple in the
sense that it consists of a single hand configuration and a single gravitational force vector, specifying a wrist
orientation. The primitives are executed open loop. Implementing all five primitives involves using the same thirty
lines of abstract Python class code with different parameters, interfacing to the robot via end-effector frames passed
to an operational space controller.

The assessment of complexity for \textit{hands} is more ambiguous. Let us compare the Shadow Hand (used in the Open AI
experiments) to the RBO Hand~3. In nearly all respects, the RBO Hand 3 is simpler (degrees of actuation, manufacturing,
material cost, control complexity, maintenance, sensing, etc.). The RBO Hand~3, however, as a result of being inherently
compliant, has a very large number of degrees of freedom; it is highly underactuated, which could be seen as complexity.
This complexity pertains to the observed behavior but does not affect the design of the rest of the system. In fact,
robust grasping~\cite{eppner_ece} and prehensile manipulation~\cite{adrian-RSS-21} have been demonstrated with simple
open-loop control, which is hard to replicate with a rigid hand like the Shadow Hand. Therefore, the RBO Hand 3 enables
simplicity in our system. 

Based on these considerations, we feel justified in saying that system we described here is very simple. We acknowledge
that this statement is not based on a clear definition of simplicity, but we hope the reader will share our intuition.

\section{Conclusion}

We presented a very simple system for robust in-hand cube reconfiguration. Our experimental results show that the
proposed system arguably exceeds the robustness and generality of competing, more complex systems. This is achieved by
combining simple GOFAI-based planning with exploiting environmental constraints to simplify control and eliminate the
need for perception. Actuation of the manipulandum is exclusively obtained from gravitational forces, varied by changing
the pose of the hand. The inherent compliance of the robot hand used in our experiments contributes robustness
by simplifying the contact dynamics between the manipulandum and the hand. We view the simplicity and performance of our
system as support for the hypothesis that GOFAI-based planning, exploitation of environmental constraints, and inherent
compliance of the end-effector form a viable conceptual basis for general-purpose manipulation. This offers an
alternative perspective to the prevalent view that deep learning alone is the most suitable approach to capable and
general manipulation.


\bibliographystyle{IEEEtranN}
\footnotesize
\balance
\bibliography{references}

\begin{thebibliography}{19}
\providecommand{\natexlab}[1]{#1}
\providecommand{\url}[1]{#1}
\csname url@samestyle\endcsname
\providecommand{\newblock}{\relax}
\providecommand{\bibinfo}[2]{#2}
\providecommand{\BIBentrySTDinterwordspacing}{\spaceskip=0pt\relax}
\providecommand{\BIBentryALTinterwordstretchfactor}{4}
\providecommand{\BIBentryALTinterwordspacing}{\spaceskip=\fontdimen2\font plus
\BIBentryALTinterwordstretchfactor\fontdimen3\font minus
  \fontdimen4\font\relax}
\providecommand{\BIBforeignlanguage}[2]{{%
\expandafter\ifx\csname l@#1\endcsname\relax
\typeout{** WARNING: IEEEtranN.bst: No hyphenation pattern has been}%
\typeout{** loaded for the language `#1'. Using the pattern for}%
\typeout{** the default language instead.}%
\else
\language=\csname l@#1\endcsname
\fi
#2}}
\providecommand{\BIBdecl}{\relax}
\BIBdecl

\bibitem[Bhatt et~al.(2021)Bhatt, Sieler, Puhlmann, and Brock]{adrian-RSS-21}
A.~Bhatt, A.~Sieler, S.~Puhlmann, and O.~Brock, ``{Surprisingly Robust In-Hand
  Manipulation: An Empirical Study},'' in \emph{Robotics: Science and Systems},
  2021.

\bibitem[Abondance et~al.(2020)Abondance, Teeple, and Wood]{wood_clark}
S.~Abondance, C.~B. Teeple, and R.~J. Wood, ``A dexterous soft robotic hand for
  delicate in-hand manipulation,'' \emph{Robotics and Automation Letters},
  vol.~5, no.~4, pp. 5502--5509, 2020.

\bibitem[Morgan et~al.(2022)Morgan, Hang, Wen, Bekris, and Dollar]{morgan_2022}
A.~S. Morgan, K.~Hang, B.~Wen, K.~Bekris, and A.~M. Dollar, ``Complex in-hand
  manipulation via compliance-enabled finger gaiting and multi-modal
  planning,'' \emph{Robotics and Automation Letters}, vol.~7, no.~2, pp.
  4821--4828, 2022.

\bibitem[Andrychowicz et~al.(2020)Andrychowicz, Baker, Chociej, Józefowicz,
  McGrew, Pachocki, Petron, Plappert, Powell, Ray, Schneider, Sidor, Tobin,
  Welinder, Weng, and Zaremba]{openai_2020}
O.~M. Andrychowicz, B.~Baker, M.~Chociej, R.~Józefowicz, B.~McGrew,
  J.~Pachocki, A.~Petron, M.~Plappert, G.~Powell, A.~Ray, J.~Schneider,
  S.~Sidor, J.~Tobin, P.~Welinder, L.~Weng, and W.~Zaremba, ``Learning
  dexterous in-hand manipulation,'' \emph{The International Journal of Robotics
  Research}, vol.~39, no.~1, pp. 3--20, 2020.

\bibitem[Chen et~al.(2023)Chen, Tippur, Wu, Kumar, Adelson, and Agrawal]{MIT}
T.~Chen, M.~Tippur, S.~Wu, V.~Kumar, E.~Adelson, and P.~Agrawal, ``Visual
  dexterity: In-hand dexterous manipulation from depth,'' in \emph{ICML
  Workshop on New Frontiers in Learning, Control, and Dynamical Systems}, 2023.

\bibitem[Pitz et~al.(2023)Pitz, Röstel, Sievers, and Bäuml]{DLR_latest}
J.~Pitz, L.~Röstel, L.~Sievers, and B.~Bäuml, ``Dextrous tactile in-hand
  manipulation using a modular reinforcement learning architecture,'' in
  \emph{IEEE International Conference on Robotics and Automation (ICRA)}, 2023,
  pp. 1852--1858.

\bibitem[Yin et~al.(2023)Yin, Huang, Qin, Chen, and Wang]{touch-dexterity}
Z.-H. Yin, B.~Huang, Y.~Qin, Q.~Chen, and X.~Wang, ``Rotating without seeing:
  Towards in-hand dexterity through touch,'' \emph{Robotics: Science and
  Systems}, 2023.

\bibitem[Khandate et~al.(2023)Khandate, Shang, Chang, Saidi, Adams, and
  Ciocarlie]{khandate2023sampling}
G.~Khandate, S.~Shang, E.~T. Chang, T.~L. Saidi, J.~Adams, and M.~Ciocarlie,
  ``Sampling-based exploration for reinforcement learning of dexterous
  manipulation,'' \emph{Robotics: Science and Systems}, 2023.

\bibitem[Kolmogorov(1965)]{kolmogorov1965three}
A.~N. Kolmogorov, ``Three approaches to the quantitative definition of
  information,'' \emph{Problems of information transmission}, vol.~1, no.~1,
  pp. 1--7, 1965.

\bibitem[Eppner et~al.(2015)Eppner, Deimel, Alvarez-Ruiz, Maertens, and
  Brock]{eppner_ece}
C.~Eppner, R.~Deimel, J.~Alvarez-Ruiz, M.~Maertens, and O.~Brock,
  ``Exploitation of environmental constraints in human and robotic grasping,''
  \emph{The International Journal of Robotics Research}, vol.~34, no.~7, pp.
  1021--1038, 2015.

\bibitem[Handa et~al.(2023)Handa, Allshire, Makoviychuk, Petrenko, Singh, Liu,
  Makoviichuk, Van~Wyk, Zhurkevich, Sundaralingam,
  et~al.]{nvidia_dextreme_2022}
A.~Handa, A.~Allshire, V.~Makoviychuk, A.~Petrenko, R.~Singh, J.~Liu,
  D.~Makoviichuk, K.~Van~Wyk, A.~Zhurkevich, B.~Sundaralingam \emph{et~al.},
  ``Dextreme: Transfer of agile in-hand manipulation from simulation to
  reality,'' in \emph{2023 IEEE International Conference on Robotics and
  Automation (ICRA)}.\hskip 1em plus 0.5em minus 0.4em\relax IEEE, 2023, pp.
  5977--5984.

\bibitem[Knight(2020)]{wired_ai}
\BIBentryALTinterwordspacing
W.~Knight, ``\BIBforeignlanguage{en-US}{{AI} {Can} {Do} {Great} {Things}—if
  {It} {Doesn}'t {Burn} the {Planet}},''
  \emph{\BIBforeignlanguage{en-US}{Wired}}, 2020. [Online]. Available:
  \url{https://www.wired.com/story/ai-great-things-burn-planet/}
\BIBentrySTDinterwordspacing

\bibitem[Dafle et~al.(2014)Dafle, Rodriguez, Paolini, Tang, Srinivasa, Erdmann,
  Mason, Lundberg, Staab, and Fuhlbrigge]{dafle_extrinsic_2014}
N.~C. Dafle, A.~Rodriguez, R.~Paolini, B.~Tang, S.~S. Srinivasa, M.~Erdmann,
  M.~T. Mason, I.~Lundberg, H.~Staab, and T.~Fuhlbrigge, ``Extrinsic dexterity:
  In-hand manipulation with external forces,'' in \emph{International
  Conference on Robotics and Automation (ICRA)}, 2014, pp. 1578--1585.

\bibitem[Chavan-Dafle and Rodriguez(2015)]{dafle_prehensile_pushing}
N.~Chavan-Dafle and A.~Rodriguez, ``Prehensile pushing: In-hand manipulation
  with push-primitives,'' in \emph{International Conference on Intelligent
  Robots and Systems (IROS)}, 2015, pp. 6215--6222.

\bibitem[Puhlmann et~al.(2022)Puhlmann, Harris, and Brock]{steffan_rh3_22}
S.~Puhlmann, J.~Harris, and O.~Brock, ``{RBO Hand 3}: A platform for soft
  dexterous manipulation,'' \emph{Transactions on Robotics}, vol.~38, no.~6,
  pp. 3434--3449, 2022.

\bibitem[Varava et~al.(2019)Varava, Welle, Mahler, Goldberg, Kragic, and
  Pokomy]{partial_caging}
A.~Varava, M.~C. Welle, J.~Mahler, K.~Goldberg, D.~Kragic, and F.~T. Pokomy,
  ``Partial caging: A clearance-based definition and deep learning,'' in
  \emph{International Conference on Intelligent Robots and Systems (IROS)},
  2019, pp. 1533--1540.

\bibitem[Erdmann and Mason(1988)]{erdmann_sensorless_1988}
M.~Erdmann and M.~Mason, ``An exploration of sensorless manipulation,''
  \emph{Journal on Robotics and Automation}, vol.~4, no.~4, pp. 369--379, 1988.

\bibitem[Erdmann(1998)]{erdmann_nonprehensile_1998}
M.~Erdmann, ``\BIBforeignlanguage{en}{An exploration of nonprehensile two-palm
  manipulation},'' \emph{\BIBforeignlanguage{en}{The International Journal of
  Robotics Research}}, vol.~17, no.~5, pp. 485--503, 1998.

\bibitem[Bai and Liu(2014)]{bai2014dexterous}
Y.~Bai and C.~K. Liu, ``Dexterous manipulation using both palm and fingers,''
  in \emph{International Conference on Robotics and Automation (ICRA)}, 2014,
  pp. 1560--1565.

\end{thebibliography}

\end{document}